\documentclass[nonacm]{acmart}

\usepackage{balance}
\usepackage{natbib}
\AtBeginDocument{%
  \providecommand\BibTeX{{%
    \normalfont B\kern-0.5em{\scshape i\kern-0.25em b}\kern-0.8em\TeX}}}

\setcopyright{rightsretained}
\copyrightyear{2024}
\acmYear{2024}

\acmDOI{XXXXXXX.XXXXXXX}

\acmConference[WebConf '24]{ACM Web Conference}{May 13--17, 2024}{Singapore}
\acmBooktitle{WebConf '24: ACM Web Conference, May 13--17, 2024, Singapore} 
\acmPrice{15.00}
\acmISBN{978-1-4503-XXXX-X/18/06}

\usepackage{hyperref}
\hypersetup{
    colorlinks=true,
    linkcolor=blue,
    filecolor=magenta,      
    urlcolor=cyan,
}

\usepackage{amsmath}
\usepackage{xcolor}
\usepackage[framemethod=tikz]{mdframed}
\usetikzlibrary{shadows}

\definecolor{mygray}{HTML}{F5F5F5}

\begin{document}

\title[FACT-GPT: Fact-Checking Augmentation via Claim Matching with LLMs]{FACT-GPT: Fact-Checking Augmentation via Claim Matching with LLMs}

\author{Eun Cheol Choi, Emilio Ferrara}
\affiliation{\institution{University of Southern California} \country{Los Angeles, CA, 90007, USA}}
\email{euncheol@usc.edu, emiliofe@usc.edu}

\renewcommand{\shortauthors}{Choi and Ferrara}

\begin{abstract}
Our society is facing rampant misinformation harming public health and trust. To address the societal challenge, we introduce FACT-GPT, a system leveraging Large Language Models (LLMs) to automate the claim matching stage of fact-checking. FACT-GPT, trained on a synthetic dataset, identifies social media content that aligns with, contradicts, or is irrelevant to previously debunked claims. Our evaluation shows that our specialized LLMs can match the accuracy of larger models in identifying related claims, closely mirroring human judgment. This research provides an automated solution for efficient claim matching, demonstrates the potential of LLMs in supporting fact-checkers, and offers valuable resources for further research in the field.

\end{abstract}





\maketitle

\section{Introduction}
The urgent need for extensive fact-checking has been driven by the rapid proliferation of misinformation on digital platforms \cite{vosoughi2018}. The fact-checking process, though complex and labor-intensive encompassing several stages from claim identification to drawing final conclusions, \cite{hassan2017, elsayed2019} could be made more efficient through AI tools \cite{arnold2020}. It is, however, critical to note that a complete automation could undermine journalistic principles and practices \cite{nakov2021automated}, thereby indicating the goal lies in enhancing, not replacing, human expertise \cite{degallier2022}.

A key element in monitoring the spread of false claims across various communication platforms is claim matching, where new instances of previously fact-checked claims are identified \cite{shaar2020}. The importance of claim matching stems from the tendency of false claims to be reused and reiterated in different formats \cite{nakov2021automated}. Effective claim matching can expedite the early detection of misinformation, content moderation, and automated debunking \cite{he2023}.

This paper explores the potential utilization of large language models (LLMs) to support the claim matching stage in the fact-checking procedure. Our study reveals that when fine-tuned appropriately, LLMs can effectively match claims. Our framework could benefit fact-checkers by minimizing redundant verification, support online platforms in content moderation, and assist researchers in the extensive analysis of misinformation from a large corpus. 

\begin{figure*}[ht]
    \centering
    \includegraphics[width=\textwidth]{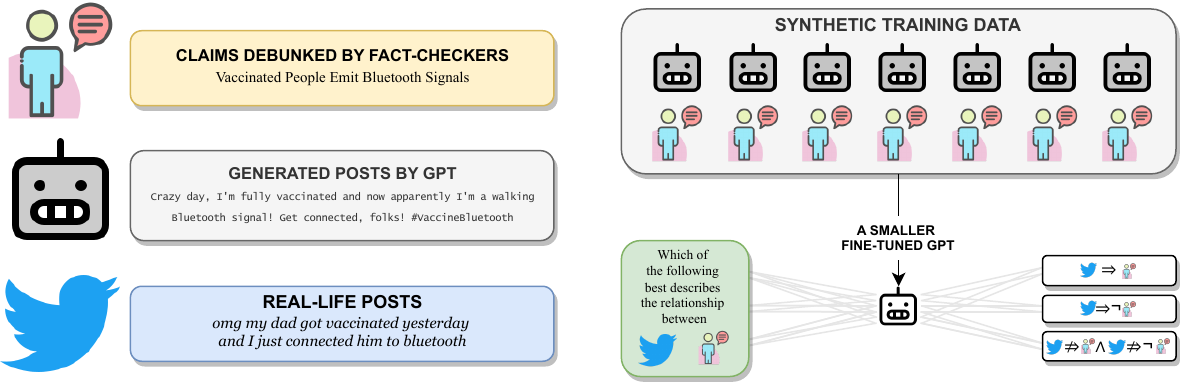}
    \caption{Overview of FACT-GPT, our framework aimed at assisting the claim matching stage of the fact-checking process}
    \label{fig:factgpt}
\end{figure*}

\section{Related Work}

\textbf{The Intersection of Fact-checkers and AI} Fact-checkers are instrumental in the fight against misinformation, as they have developed reliable practices and principles over time \cite{ifcn2023}. The integration of AI into the fact-checking process should be conducted with great care, with the goal of enhancing efficiency without undermining established principles \cite{nakov2021automated}. AI models that support rather than replace fact-checkers are more likely to be embraced. While fact-checkers have shown interest in AI tools for identifying claims and assessing their virality \cite{arnold2020}, they maintain skepticism about AI entirely replacing human intervention, thereby highlighting the indispensable role of human judgment.

\textbf{LLMs in Annotation Tasks} Large Language Models (LLMs) have garnered significant interest due to their potential to automate diverse annotation tasks. Despite platforms like Amazon Mechanical Turk (MTurk) enabling crowd-sourced annotation, creating comprehensive datasets for complex tasks continues to be expensive. Given their flexible nature, LLMs' performance in various annotation tasks is being scrutinized. Research has evaluated LLMs in contexts such as fact-checking \cite{hoes2023}, annotating tweets \cite{gilardi2023}, and beyond. Generating synthetic training data to enhance LLMs' performance in classification tasks has also been explored \cite{dai2023}. However, it is crucial to acknowledge LLMs' inherent limitations. Their probabilistic nature implies that their outputs can vary according to prompts and parameters \cite{reiss2023}. When compared to task-specific models, ChatGPT often underperform \cite{kocon2023, zhu2023}, underlining the need for models that are specifically designed and utilized for certain tasks.

\section{Proposed Framework}

\subsection{Task Description} To evaluate various Large Language Models' (LLMs) performance in claim matching, we employ a \textit{textual entailment task}. Textual entailment involves categorizing relationships between pairs of statements into three unique classes: \textit{Entailment}, \textit{Neutral}, and \textit{Contradiction}. A pair is classified as \textit{Entailment} if the veracity of Statement A inherently implies the truth of Statement B. The pair is labeled as \textit{Neutral} if the truthfulness of statement A doesn't affirm or deny statement B's truth. It's identified as \textit{Contradiction} if Statement A's truth infers that Statement B is false. Textual entailment tasks are centered around everyday reasoning rather than strict logic, hence human judgment and common sense establish the ground truth \cite{marelli2014, pado2022}. This kind of task has previously showed effectiveness in detecting rumors \cite{yavary2019}.

Claim matching tasks can be configured in various forms including but not limited to textual entailment \cite{ma2019}, ranking \cite{shaar2022, lagatta2023}, and binary detection tasks \cite{jin2022}. Defining claim matching as a 3-class entailment task poses both advantages and challenges. Identifying contradicting pairs is important as such rebuttals play a crucial role in mitigating the spread of misinformation \cite{he2023, tambuscio2019}. However, it's challenging due to the scarcity of contradiction pairs in real-world instances \cite{marelli2014}.

\subsection{Datasets}
In this study, we focus on misinformation relating to public health, specifically COVID-19 related false claims that have been fact-checked. 1,225 False claims debunked by professional fact checkers in 2020 and 2021 were obtained from \href{https://toolbox.google.com/factcheck/explorer}{\textcolor{blue}{\textit{Google Fact Check Tools}}} and \href{https://www.politifact.com/}{\textcolor{blue}{\textit{PolitiFact}}}.

\begin{figure}[b]
    \centering
    \begin{tikzpicture}[font=\footnotesize]
        \node[draw, fill=mygray, rounded corners, drop shadow={fill=black!30, shadow xshift=3pt, shadow yshift=-3pt, opacity=0.5}, inner sep=10pt] {
            \begin{minipage}{\columnwidth}
                \begin{tabular}{l p{\columnwidth}}
                    \texttt{System} & Generate TWEET so that if TWEET is true, then CLAIM is also true. Be brief. Do not start a sentence with 'Just'. \\
                     & \\
                    \texttt{Input} & Vaccininated people emit Bluetooth signals. \\
                     & \\
                    \texttt{Output} & Crazy day. I'm fully vaccinated and now apparently I'm a walking Bluetooth signal! Get connected, folks! \#VaccineBluetooth \\
		      \end{tabular}
            \end{minipage}
            };
    \end{tikzpicture}

    \caption{Example of synthetic tweet generation prompts}
    \label{fig:gen_prompt}
\end{figure}

\subsubsection{Synthetic Training Datasets Generation}\label{trainset}
We utilized Large Language Models (LLMs) to generate synthetic training data, allowing for the creation of a balanced dataset specifically designed for claim matching tasks. Fine-tuning language models on synthetic datasets can enhance their adaptability to specific task nuances, potentially leading to better classification accuracy. In addition, fine-tuning smaller models reduces the computational cost involved in large-scale operations while making it easier to customize these models based on emerging new claims.

To generate synthetic training data, we utilized three language models available via the \href{https://openai.com/product}{\textcolor{blue}{\textit{OpenAI API}}} or the \href{https://huggingface.co/docs/api-inference/en/index}{\textcolor{blue}{\textit{HuggingFace Inference API}}}: \texttt{GPT-4}, \texttt{GPT-3.5-Turbo}, and \texttt{Llama-2-70b-chat-hf}. Using a collection of debunked claims as a basis, we generated tweets that either supported, were neutral to, or contradicted these claims. To generate varied styles in the outputs by the language models, we set the temperature parameter at 1. Figure \ref{fig:gen_prompt} provides an example of a prompt used for data generation. A total of 3,675 synthetic tweets were generated from each model, ensuring an equal distribution across all three categories.

\subsubsection{Ground Truth Dataset}

\begin{figure}[b]
    \centering
    \includegraphics[width=0.45\textwidth]{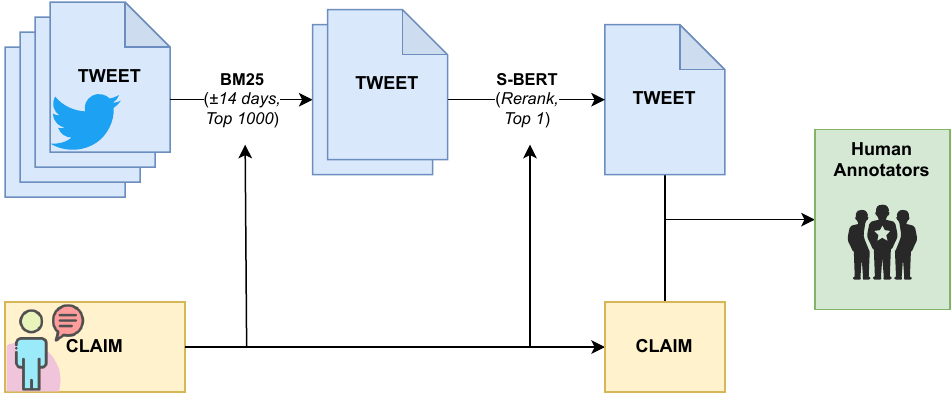}
    \caption{Workflow of test data construction}
    \label{fig:workflow}
\end{figure}

Our method for creating a ground truth dataset is illustrated in Figure \ref{fig:workflow}. Initially, we paired tweets from the publicly available \href{https://github.com/echen102/COVID-19-TweetIDs}{\textcolor{blue}{\textit{Coronavirus Twitter Dataset}}} \cite{chen2020} with debunked false claims, considering both token and semantic similarities. This process generated a unique set of 1,225 pairs consisting of tweets and claims. Experienced annotators on Amazon Mechanical Turk then classed each of these pairs into one of the three categories. The final categorization was based on which class received the majority of votes, creating a fully annotated test dataset, as illustrated in Table \ref{tab:freq}.

\begin{table}[t]
\small
  \caption{Descriptive statistics for test data.}
  \label{tab:freq}
  \begin{tabular}{lcc}
    \toprule
    Label & Count & Percentage  \\
    \midrule
    ENTAILMENT     & 647 & 52.8\% \\
    NEUTRAL        & 433 & 35.3\% \\
    CONTRADICTION  &  90 & 7.3\% \\
    (Two-way ties)  &  55 & 4.5\% \\
    \midrule
    TOTAL          & 1225 & 100\% \\
    \bottomrule
  \end{tabular}
\end{table}

\begin{figure}[h]
    \centering
    \begin{tikzpicture}[font=\footnotesize]
        \node[draw, fill=mygray, rounded corners, drop shadow={fill=black!30, shadow xshift=3pt, shadow yshift=-3pt, opacity=0.5}, inner sep=10pt] {
            \begin{minipage}{\textwidth}
                \begin{tabular}{l p{\textwidth}}
                    \texttt{System} & Which of the following best describes the relationship between TWEET and CLAIM? \\
                    & You must choose from ENTAILMENT, NEUTRAL, or CONTRADICTION. \\
                    & \\
                    & If TWEET is true: \\
                    & (ENTAILMENT) then CLAIM is also true. \\
                    & (NEUTRAL) CLAIM cannot be said to be true or false. \\
                    & (CONTRADICTION) then CLAIM is false. \\
                    & \\
                    \texttt{Input} &  TWEET: omg my dad got vaccinated yesterday and I just connected him to bluetooth \\
                    & CLAIM: Vaccininated people emit Bluetooth signals. \\
                    & \\
                    \texttt{Output} & ENTAILMENT \\
                \end{tabular}
            \end{minipage}
        };
    \end{tikzpicture}
    \caption{Example of an entailment task prompt}
    \label{fig:cot}
\end{figure}

\subsection{Experiments}

\subsubsection{Baselines} We established comparison benchmarks by assessing the performance of several pre-trained Large Language Models (LLMs), including \texttt{GPT-4}, \texttt{GPT-3.5-Turbo}, \texttt{Llama-2-13b}, and \texttt{Llama-2-7b}, against human annotations. We adjusted the temperature setting to 0 (or 0.01 for Llama models) to make the annotation process as consistent as possible. We then presented entailment task prompts to each LLM and collected their responses.
 
\subsubsection{Fine-tuning} 
Our assessment of FACT-GPT's effectiveness involved fine-tuning \texttt{GPT-3.5-Turbo}, \texttt{Llama-2-13b}, and \texttt{Llama-2-7b} with the synthetic training dataset outlined in \ref{trainset}. We allocated 80\% of the data for training and the remaining 20\% for validation. \texttt{GPT-3.5-Turbo} underwent fine-tuning using \href{https://platform.openai.com/docs/guides/fine-tuning}{\textcolor{blue}{\textit{\textit{OpenAI's Fine-tuning API}}}}. Meanwhile, for the LLaMa models, we applied LoRA (Low-Rank Adaptation, \cite{hu2021}) in \href{https://github.com/hiyouga/LLaMA-Factory}{\textcolor{blue}{\textit{\textit{LLaMa-Factory}}}} \cite{llama-efficient-tuning}, which is an efficient tuning framework for LLMs. \texttt{BERT-base} model was fine-tuned on \texttt{GPT-4}-generated train set to provide an additional benchmark. Each model went through three epochs (five for \texttt{BERT-base}) of fine-tuning on a single A100 GPU.

\subsubsection{Results} The overall performance of FACT-GPTs are summarized in Table \ref{tab:performance}. Notably, models fine-tuned on synthetic datasets exhibited superior performance in comparison to the pre-trained versions. There was a consistent pattern in the performance among the fine-tuned models, with all models exhibiting improved outcomes when fine-tuned using training data generated by \texttt{GPT-4} as opposed to those generated by \texttt{GPT-3.5-Turbo} or \texttt{Llama-2-70b}. This trend emphasizes the significance of the quality of training data in determining the effectiveness of the resulting models.

Table \ref{tab:labelperformance} reveals that our top-performing models are more adept at classifying \textit{Entailment} and \textit{Neutral} labels, but face challenges with \textit{Contradiction} labels. This suggests that our FACT-GPTs are proficient in determining the relevance or irrelevance of social media posts to the original debunked claims. However, given that rebuttals to false claims play a crucial role in preventing the spread of misinformation \cite{he2023, tambuscio2019}, future work should focus on improving the detection of contradictory posts.

\begin{table}[t]
\small
  \caption{Overall performance of pre-trained and fine-tuned models.}
  \label{tab:performance}
  \begin{tabular}{llccc}
    \toprule
    Model & Train Set From & Precison & Recall & Accuracy  \\
    \midrule
    \texttt{BERT-base}        & \textit{GPT-4} & .46 & .46 & .46 \\
    \midrule
    \texttt{GPT-4}        & --- & .64 & .70 & .63 \\
    \midrule    
    \texttt{GPT-3.5-Turbo} & \textit{GPT-4}                 &        .64 &     .68 &       .73 \\
                     & \textit{GPT-3.5-Turbo}         &        .51 &     .59 &       .57 \\
                     & \textit{Llama-2-70b}           &        .57 &     .65 &       .60 \\
                     & --- & .56 & .61 & .58 \\
    \midrule
    \texttt{Llama-2-13b}   
          & \textit{GPT-4}                 &        .63 &     .69 &       .71 \\
                     & \textit{GPT-3.5-Turbo}         &        .52 &     .57 &       .60 \\
                     & \textit{Llama-2-70b}           &        .56 &     .64 &       .63 \\
                     & --- & .51 & .47 & .30 \\
    \midrule
    \texttt{Llama-2-7b}    
          & \textit{GPT-4}                 &        .64 &     .70 &       .73 \\
                     & \textit{GPT-3.5-Turbo}         &        .48 &     .52 &       .56 \\
                     & \textit{Llama-2-70b}           &        .60 &     .60 &       .68 \\
                     & --- & .42 & .46 & .40 \\
    \bottomrule
  \end{tabular}
\end{table}

\begin{table}[t]
\small
  \caption{Label-by-label performance of pre-trained and fine-tuned models.}
  \label{tab:labelperformance}
  \begin{tabular}{llccc}
    \toprule
    Model & Train Set From & $F1_{Ent}$ & $F1_{Neu}$ & $F1_{Con}$  \\
    \midrule
    \texttt{BERT-base}        & \textit{GPT-4} & .65 & .30 & .21 \\
    \midrule
    \texttt{GPT-4}        & --- & .61 & .63 & .51 \\
    \midrule    
    \texttt{GPT-3.5-Turbo} 
        & \textit{GPT-4}                 &        .83 &     .67 &       .44 \\
                     & \textit{GPT-3.5-Turbo}         &        .76 &     .35 &       .32 \\
                     & \textit{Llama-2-70b}           &        .73 &     .57 &       .34 \\
                     & --- & .58 & .64 & .39 \\
    \midrule
    \texttt{Llama-2-13b}   
          & \textit{GPT-4}                 &        .79 &     .69 &       .45 \\
                     & \textit{GPT-3.5-Turbo}         &        .73 &     .50 &       .34 \\
                     & \textit{Llama-2-70b}           &        .74 &     .57 &       .39 \\
                     & --- & .37 & .36 & .19 \\
    \midrule
    \texttt{Llama-2-7b}    
          & \textit{GPT-4}                 &        .79 &     .72 &       .46 \\
                     & \textit{GPT-3.5-Turbo}         &        .69 &     .41 &       .29 \\
                     & \textit{Llama-2-70b}           &        .74 &     .65 &       .40 \\
                     & --- & .63 & .02 & .26 \\
    \bottomrule
  \end{tabular}
\end{table}

\section{Discussion}
This study underscores the potential of large language models (LLMs) in augmenting the fact-checking process, particularly during the claim matching phase. Our research demonstrates that LLMs have the capacity to discern entailment relationships between social media posts and debunked claims. Importantly, our study reveals that appropriately fine-tuned, smaller LLMs can yield a performance comparable to larger models, thereby offering a more accessible and cost-effective AI solution without compromising quality. However, while our models excel in detecting whether social media content is relevant to or irrelevant from debunked claims, they show struggles with categorizing posts that contradict these claims.  This is an area that requires further refinement, given the importance of rebuttals in curbing the spread of misinformation. 

Looking forward, it is crucial to encourage ongoing collaborations among researchers, developers, and fact-checkers to fully exploit AI benefits while mitigating its potential drawbacks. The importance of human expertise and supervision in this context cannot be overstated. Completely automating fact-checking procedures using AI carries certain risks and limitations, such as the perpetuation of biases intrinsic to models and inherent inconsistencies due to their probabilistic nature. However, with thoughtful incorporation, technologies could substantially augment the capabilities of fact-checkers to detect and debunk misinformation. 

Future studies should focus on discovering different methods for data synthesis and augmentation to further optimize FACT-GPT. Additionally, evaluating the model's performance across a variety of real-world datasets is crucial. Exploration into the integration of natural language explanation (NLE) capabilities within GPT models can further enhance transparency. This research adds substantively to a growing body of work examining the use of LLMs in support of human fact-checkers, offering a foundation for continued studies and the responsible advancement of AI tools to effectively combat the spread of misinformation at a larger scale.

\smallskip \small \textbf{Acknoledgements}. This work was supported in part by DARPA (contract no. HR001121C0169).

\balance
\bibliographystyle{abbrvnat}

\bibliography{sample-base}


\end{document}